# Remote Manipulation of Multiple Objects with Airflow Field Using Model-Based Learning Control

Artur Kopitca, Shahriar Haeri, Quan Zhou

*Abstract*—Non-contact manipulation is an emerging and highly promising methodology in robotics, offering a wide range of scientific and industrial applications. Among the proposed approaches, airflow stands out for its ability to project across considerable distances and its flexibility in actuating objects of varying materials, sizes, and shapes. However, predicting airflow fields at a distance – as well as the motion of objects within them – remains notoriously challenging due to their nonlinear and stochastic nature. Here, we propose a model-based learning approach using a jet-induced airflow field for remote multi-object manipulation on a surface. Our approach incorporates an analytical model of the field, learned object dynamics, and a model-based controller. The model predicts an air velocity field over an infinite surface for a specified jet orientation, while the object dynamics are learned through a robust system identification algorithm. Using the model-based controller, we can automatically and remotely, at meter-scale distances, control the motion of single and multiple objects for different tasks, such as path-following, aggregating, and sorting.

## I. INTRODUCTION

Airflow possesses a remarkable ability to actuate objects without physical contact over considerable distances. While other approaches to non-contact manipulation—such as those based on magnetic [1], acoustic [2], and optical [3] stimuli—are often limited to milli- or microscale particles or specific material properties, airflow can actuate objects across a wide spectrum of sizes and materials, as observed in nature. Consequently, it has been employed in systems such as air-bearing tables for precise wafer transport [4] and noncontact grippers for fabrics [8], as well as in systems without individual object control, such as stationary and mobile robotic blowers to unfold clothes [5] and sweep leaves [6]. Using the Coandă effect, airflows can trap and move specifically shaped objects such as spheres and cylinders [7]. Additionally, impinging air jets are employed to blow microparticles [9] and droplets [10] from surfaces. Despite the impressive achievements, all these airflow-based techniques are either restricted to specific object shapes or unable to control the motion of individual objects.

Recently, the authors have introduced a highly promising method for automatic, non-contact manipulation of diverse objects at meter-scale distances on a surface using a jet-induced airflow field [11]. This method enables precise path-following of cm-sized objects with a mean precision of 1.5 cm, at distances up to 2.7 meters away from the jet source. The method has demonstrated its versatility with various object materials and shapes, including both regular and irregular geometries, as well as deformable objects. Additionally, this method shows potential across different environments and applications, such as manipulating tethered agents to retrieve heavy objects and maneuvering untethered soft agents to close an electrical circuit.

However, predicting such jet-induced airflow fields and the resulting motion of objects on a surface remains a significant challenge. Conventional techniques for simulating or measuring airflows, such as Computational Fluid Dynamics (CFD) [12] and Particle Image Velocimetry (PIV) [13], are computationally intensive or impractical for real-time predictions. Additionally, nonlinear [14] and stochastic nature of object motion within airflow fields adds further complexity to aerodynamic predictions. These challenges prevent the application of advanced control algorithms to complex manipulation tasks, such as multi-object manipulation.

In this paper, we show that a jet-induced airflow field on a surface can be analytically modelled, and the resultant object motion can be learned. Hence, we can achieve model-based learning control for remote manipulation of multiple objects using the airflow field. The proposed analytical model predicts an air velocity field over an infinite plane surface for a specified jet orientation. By combining this model with a robust dynamic system identification algorithm, we can learn the dynamics of objects within the field and manipulate them using a model-based controller. Our approach unlocks new possibilities for airflow-field-based manipulation, including the simultaneous manipulation of multiple objects. Specifically, in addition to single-object manipulation, we can automatically steer multiple objects along reference paths, aggregate scattered objects into reference zones, and sort objects into groups.

The rest of the article is organized as follows. Section II overviews the setup and operational principle of remote airflow-field-based manipulation. Section III presents the analytical model of the airflow field. Section IV covers the dynamic system identification algorithm and its results. Section V outlines the model-based controller and demonstrates single- and multi-object manipulation. Section VI concludes the article.

This work was supported by the Aalto Doctoral School of Electrical Engineering. *(Corresponding author: Quan Zhou).*

The authors are with the Department of Electrical Engineering and Automation of Aalto University, Maarintie 8, 02150 Espoo, Finland (e-mail: artur.kopitca@aalto.fi; shahriar.haeri@aalto.fi; quan.zhou@aalto.fi).

This article has supplementary material provided by the authors.

## II. SYSTEM OVERVIEW

### A. Remote Manipulation Principle and Setup

The non-contact manipulation is conducted using a remotely generated airflow field, as illustrated in Fig. 1. To generate the airflow field, we project an air jet onto a plane floor surface from a meter-scale distance, using the setup shown in Fig. 1(a). A jet was chosen for its simplicity, with the future possibility of adding additional jets. To adjust the airflow field, we rotate the converging air nozzle (SMC KN-R02-250, outlet diameter 2.5 mm) emitting the jet in the pan $\theta \in [-180°, 180°]$ and tilt $\varphi \in [0°, 90°]$ angular directions using two servomotors (Dynamixel XM430-W210-T).

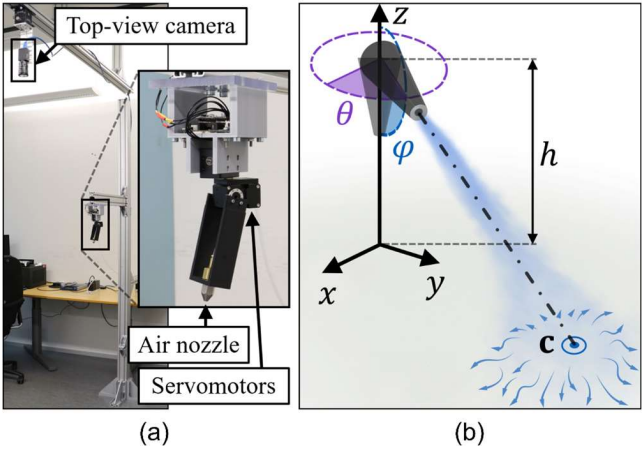

Fig. 1. (a) Photograph of the manipulation setup. (b) Schematic of an air jet projected onto a surface and the jet-induced airflow field (blue arrows); parameters: nozzle pan angle $\theta$, nozzle tilt angle $\varphi$, projection point $\mathbf{c}$, and distance $h$ from the tilting servo shaft to the surface.

The projection point, i.e., the point on the surface at which the nozzle is directed, denoted by $\mathbf{c} = [x_c, y_c]^T$, is shown in Fig. 1(b). The projection point can be computed from the pan $\theta$ and tilt $\varphi$ angles of the nozzle as follows,

$$\mathbf{c} = h \tan \varphi \begin{bmatrix} \cos \theta \\ \sin \theta \end{bmatrix} \quad (1)$$

where $h$, set to 1.43 m, is the distance from the tilting servo shaft to the surface, as detailed in Fig. 1(b), and the origin ($x = 0, y = 0$) is assumed to be directly beneath the nozzle on the floor surface. The parameters, including $h$, were previously tuned to increase the feasible manipulation region of tracer objects [11], which were also used throughout most of this work (see Section II-B).

Additionally, we employed a top-view camera (FLIR Grasshopper3 GS3-U3-41C6M with Edmund Optics 12 mm fixed focal length HP series lens) to acquire images for visual feedback, as shown in Fig. 1(a). The camera operated at 1024×1024 pixels and 40 Hz.

### B. Objects for Manipulation

As proof of concept, we used tracer objects (polystyrene hemispheres, ø 3 cm, 0.16 g) for most of the modelling and manipulation experiments, unless stated otherwise.

## III. MODELLING THE JET-INDUCED AIRFLOW FIELD

To predict the air velocity $\mathbf{v}_{\text{air}} \in \mathbb{R}^2$ at any point on the surface for a given air nozzle orientation $(\theta, \varphi)$ in real time, we require an analytical airflow field model, as simulations [12] and measurements [13] are impractically time-consuming for this purpose. To collect the data for this model, we measured air velocity and performed CFD simulations across a 2 m × 2 m 1.5 cm workspace, but limited our data collection to only three tilt angles, $\varphi = [0, 22.5, 45]°$. Due to the rotational symmetry of the system, we held the pan angle fixed at $\theta = 90°$. The height of 1.5 cm above the floor surface corresponds to the size of the tracer objects used (see Section II-B). More details on data collection are provided in Appendices A and B.

### A. Airflow Field Model

The airflow field model we aim to build should capture key properties of the jet-induced airflow field.

One key property is that, in addition to the projection point $\mathbf{c}$, the airflow field has a stagnation point, denoted by $\mathbf{s} = [x_s, y_s]^T$, where the air velocity reaches zero and static pressure peaks [11], [12], following Bernoulli's principle. Using our CFD data, we can locate the stagnation points for the three nozzle tilt angles $\varphi$, as shown in Fig. 2(a). For $\varphi = 0°$, the stagnation and projection points coincide, and as $\varphi$ increases, the distance between these two points grows exponentially, which can be described by the following relationship,

$$\|\mathbf{c} - \mathbf{s}\| = a_1 \|\mathbf{c}\|^{a_2} \quad (2)$$

where $a_1$ and $a_2$ are model coefficients. Based on the CFD data, we obtained $a_1 = 0.1$ and $a_2 = 2.33$, as plotted in Fig. 2(b). Using this simple relationship (2), we can compute the stagnation point from the projection point.

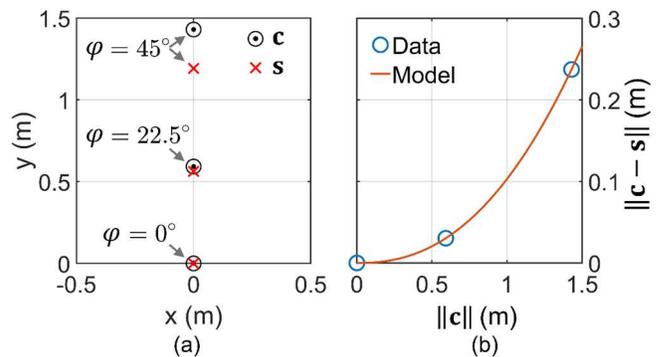

Fig. 2. (a) Projection points $\mathbf{c}$ and stagnation points $\mathbf{s}$ at different nozzle tilt angles $\varphi$. (b) Modelled relationship between $\mathbf{c}$ and its distance to $\mathbf{s}$.

Another well-documented property is that air velocity magnitude increases from zero at the stagnation point to a peak value before decreasing again [15]. To capture such phenomena, we employed a two-term exponential equation,

$$v_{\text{air}}(r) = b_1(e^{b_2 r} - e^{b_3 r}) \quad (3)$$

where $v_{\text{air}} = \|\mathbf{v}_{\text{air}}\|$ is the air velocity magnitude, $r$ is the radial distance from the stagnation point $\mathbf{s}$, and $b_i$ are fitted coefficients. These coefficients vary with both the tilt angle ($\varphi$) and the radial direction ($\alpha$) from $\mathbf{s}$, reflecting the asymmetry of the airflow [11] (see Fig. 3(a) for examples of the radial directions $\alpha$). Thus, we fitted the coefficients individually for each combination of $\alpha$ (measured at one-degree intervals) and $\varphi$. Fitting was performed using a weighted combination of measurements and CFD data (see Appendix C for more details).

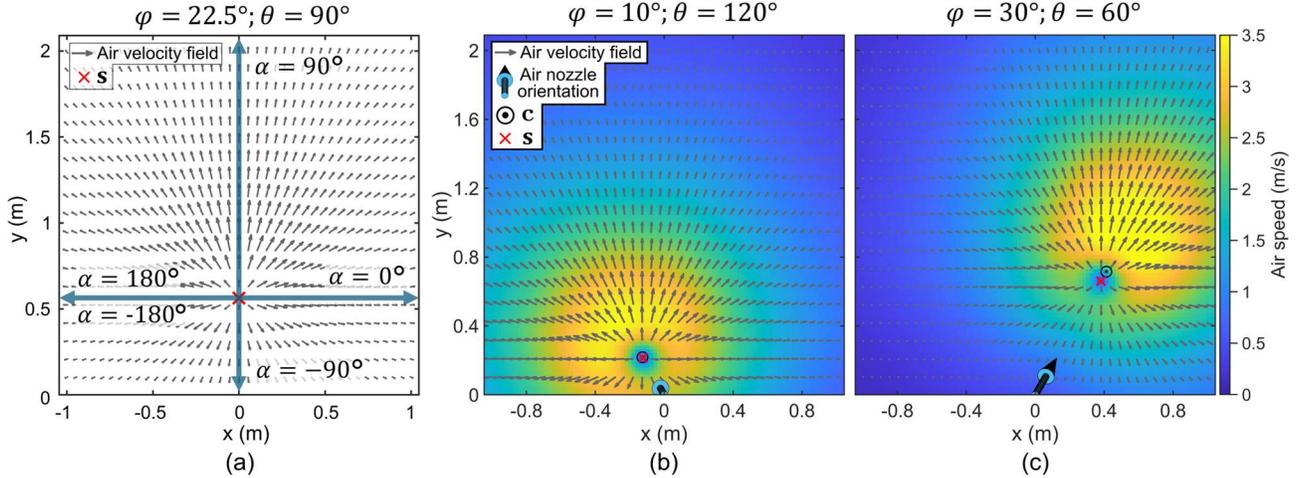

Fig. 3. (a) Air velocity field based on the CFD data, overlaid with various radial directions ($\alpha$) originating from the stagnation point ($\mathbf{s}$). (b), (c) Air velocity fields predicted by the airflow field model at different air nozzle orientations $(\theta, \varphi)$. The projection point is denoted by $\mathbf{c}$.

With this simple model, we can predict the air velocity at any point for a given air nozzle orientation by following three steps: (i) calculate the projection (1) and stagnation (2) points, (ii) determine the radial distance and direction from the stagnation point, and (iii) compute the air velocity magnitude (3) using the fitted coefficients. For values beyond the fitted range, 1D linear interpolation is applied for $\alpha$, and Makima interpolation [16] for $\varphi$. The air velocity direction is assumed to align with $\alpha$, i.e., pointing away from the stagnation point, an assumption supported by the CFD data. Examples of predicted air velocity fields for various air nozzle orientations are shown in Fig. 3(b)-(c) and the supplementary video.

The proposed airflow field model is computationally efficient, relying on simple arithmetic operations and 1D interpolations, allowing for real-time predictions. The model has a time complexity of $O(N)$, where $N = 3$ is the number of $\varphi$ values used for interpolation. However, caution should be taken when predicting beyond the range of $\varphi$ values used to fit the model, as extrapolation may lead to significant errors.

*B. Model Accuracy*

The model agrees well with both the CFD data and measurements, achieving mean absolute percentage errors (MAPEs) between 8.2% and 18% in most cases. One exception occurs at $\varphi = 45°$, where the MAPE between the model and the CFD data is 58.1%. This discrepancy is likely due to limitations of the hot-wire anemometer used in the measurements that the model was fitted to (see Appendix A). Specifically, the hot-wire anemometer may overestimate low air velocities near its lower detection limit (~10 cm/s), which typically occur at distances greater than ~80 cm from the stagnation point. Future work could improve accuracy by utilizing more advanced measurement techniques, such as PIV [13].

IV. MODELLING THE MOTION OF OBJECTS

To model the motion of objects in airflow, we first review the physics governing such motion. An object in a jet-induced airflow field experiences aerodynamic forces such as drag and pressure gradient forces, opposed by friction [11]. According to Newton's second law, the translational 2D motion of the object on the surface follows

$$m_{\text{obj}} \frac{d\mathbf{v}_{\text{obj}}}{dt} = \mathbf{F}_{\text{D}} + \mathbf{F}_{\text{PG}} - \mathbf{F}_{\text{f}} \quad (4)$$

where $m_{\text{obj}}$ is the object mass, $\mathbf{v}_{\text{obj}}$ its 2D velocity, $\mathbf{F}_{\text{D}}$ the drag force, $\mathbf{F}_{\text{PG}}$ the pressure gradient force, and $\mathbf{F}_{\text{f}}$ the dry friction force.

Among the three forces, the drag force, $\mathbf{F}_{\text{D}}$, primarily depends on the local air velocity $\mathbf{v}_{\text{air}}$ relative to the object velocity $\mathbf{v}_{\text{obj}}$ and the drag coefficient. The latter is influenced by the Reynolds number, which itself depends on the relative air velocity [17]. Other factors affecting drag, such as the object shape and the properties of the air, are assumed to be constant.

The pressure gradient force, $\mathbf{F}_{\text{PG}}$, is significant near the stagnation point $\mathbf{s}$, where the pressure peaks and dissipates with distance [11], [12]. Similarly, object motion is most stochastic near this point [11]. To mitigate the stochastic effects, we control the object motion only at a sufficient distance from $\mathbf{s}$ (see Section V), where $\mathbf{F}_{\text{PG}}$ is assumed to be negligible.

The dry friction force, $\mathbf{F}_{\text{f}}$, can vary with the object velocity relative to the contacting surface [18], which reduces to $\mathbf{v}_{\text{obj}}$ due to the stationary surface.

Thus, the model of the object motion in (4) simplifies to an ordinary differential equation (ODE) of the form

$$\frac{d\mathbf{v}_{\text{obj}}}{dt} = f(\mathbf{v}_{\text{obj}}, \mathbf{v}_{\text{air}}) \quad (5)$$

where $f$ is a function combining the object mass and forces affecting velocity change, which we aim to identify.

Note that the object motion is also stochastic due to turbulence [11] and transient effects resulting from the jet changing direction. These effects are addressed in the system identification process through additional measures to enhance noise robustness (see Section IV-A).

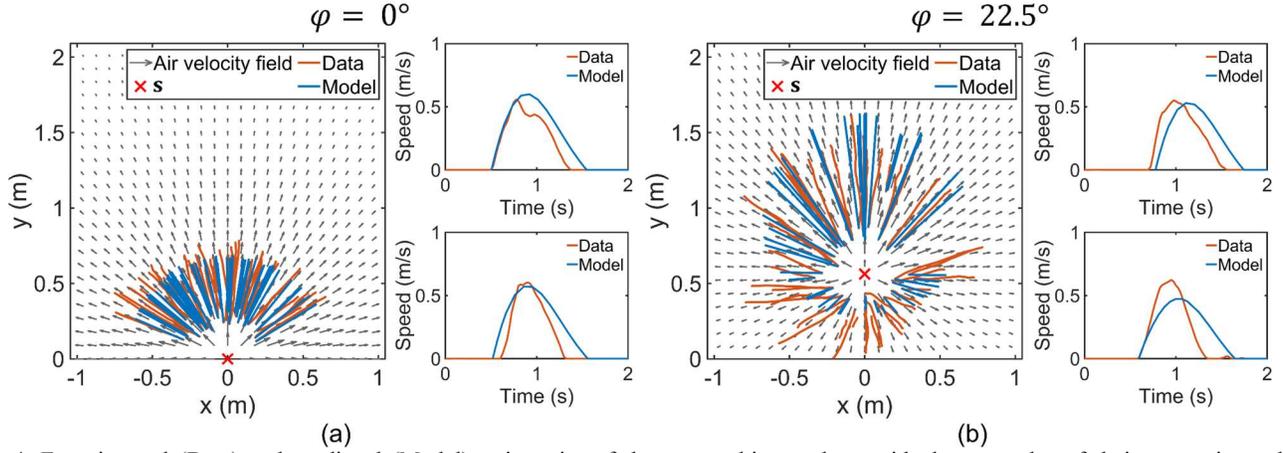

Fig. 4. Experimental (Data) and predicted (Model) trajectories of the tracer objects, along with the examples of their respective velocity magnitudes (Speed), at (a) $\varphi = 0°$ and (b) $\varphi = 22.5°$ tilt angles of the air nozzle. The stagnation points are denoted by **s**.

*A. Robust Dynamic System Identification*

The object motion model in (5) is nonlinear and has an unknown structure and parameters. We propose using the Sparse Identification of Nonlinear Dynamics (SINDy) algorithm [19] to identify both the model structure and its parameters. Given the existence of numerous drag [17] and friction [18] models of varying complexity, SINDy provides a means to identify a force model that is both accurate and sparse, meaning it uses only the most essential terms. This identified model will enable us to predict the translational 2D motion of tracer objects within the jet-induced airflow field.

The SINDy algorithm we employed identifies the dynamics in (5) by first expressing it in terms of the following discrete-time data matrices,

$$\mathbf{V}_{\text{obj}}^{(k+1)} = \Theta(\mathbf{V}_{\text{obj}}^{k}, \mathbf{V}_{\text{air}})\Xi \qquad (6)$$

where $\mathbf{V}_{\text{obj}}^{k} = [v_{\text{obj}}^{1}, \ldots, v_{\text{obj}}^{k}]^{\text{T}} \in \mathbb{R}^{k \times 1}$ and $\mathbf{V}_{\text{obj}}^{(k+1)} = [v_{\text{obj}}^{2}, \ldots, v_{\text{obj}}^{(k+1)}]^{\text{T}} \in \mathbb{R}^{k \times 1}$ each contain $k$ snapshots of the magnitudes of the object velocity. Here, $v_{\text{obj}}^{k} = \|\mathbf{v}_{\text{obj}}^{k}\|$ is the magnitude of the object velocity at timestep $k$. $\mathbf{V}_{\text{air}} = [v_{\text{air}}^{1}, \ldots, v_{\text{air}}^{k}]^{\text{T}} \in \mathbb{R}^{k \times 1}$ contains $k$ air velocity magnitudes at the object position at each timestep, calculated using the airflow field model (Section III). The function $\Theta(\mathbf{V}_{\text{obj}}^{k}, \mathbf{V}_{\text{air}}) \in \mathbb{R}^{k \times m}$ is a library of $m$ candidate functions, and $\Xi \in \mathbb{R}^{m \times 1}$ contains $m$ unknown coefficients to be fitted.

The conversion of the continuous-time system (5) into the discrete-time format (6) avoids the challenge of estimating the object acceleration, $\frac{dv_{\text{obj}}}{dt}$, which can be noisy due to turbulence. Instead, the discrete-time system provides a cleaner path to recover the continuous-time dynamics using a forward difference approximation.

For the library $\Theta$, we included a constant term and all possible product combinations of $\mathbf{V}_{\text{obj}}^{k}$ and $\mathbf{V}_{\text{air}}$ up to the third order. These terms were selected to capture possible nonlinearities, which may appear in drag [17] and friction [18]. Although higher-order polynomials and trigonometric terms could be added, they often lead to overfitting; thus, starting with a low-order polynomial basis is a recommended approach [19].

Hence, the system identification task is posed as a sparse regression problem,

$$\Xi = \arg\min_{\widehat{\Xi}} \left\| \mathbf{V}_{\text{obj}}^{(k+1)} - \Theta(\mathbf{V}_{\text{obj}}^{k}, \mathbf{V}_{\text{air}})\widehat{\Xi} \right\|^{2} + \lambda \|\widehat{\Xi}\|_{1} \qquad (7)$$

where $\|\cdot\|_{1}$ is a $L^{1}$-norm and $\lambda$, set to 0.07, is a regularization parameter promoting sparsity. In other words, we aim to identify the active terms in $\Theta$ that define the object dynamics, thus determining the model that is both accurate and sparse.

While the classical SINDy algorithm [19] employs sequentially thresholded least-squares (STLS) to solve (7), its sensitivity to noise led us to use two modifications: Ensemble-SINDy [20] and robust regression [21]. Ensemble-SINDy identifies an ensemble of models from random bootstraps of the data and aggregates the results by taking the median of the coefficients. We used 15 bootstraps in our implementation. Robust regression, on the other hand, mitigates the effect of outliers by applying the bisquare function to the residuals, reducing the influence of noise during system identification.

*B. Results of System Identification for Tracer Objects*

To conduct the identification using SINDy, we collected trajectories of the tracer objects in the jet-induced airflow field at two nozzle tilt angles: 0° and 22.5° (details on the data collection can be found in Appendix D). We chose these two angles because they cover a wide range of air velocity magnitudes, up to 3.4 m/s, and their behaviors. By applying SINDy, the following 1D system was identified,

$$\frac{dv_{\text{obj}}}{dt} = \xi_{1} v_{\text{obj}} + \xi_{2} v_{\text{air}} + \xi_{3} \qquad (8)$$

where the fitted coefficients are $\xi_{1} = -2.66$, $\xi_{2} = 4.26$, and $\xi_{3} = -8.53$; the goodness of fit was $R^{2} = 0.95$. It is interesting to note that the resultant model structure is simple, which is an elegant result. Nevertheless, although the identified system (8) is linear in the coefficients $\xi_{i}$, the air velocity magnitude $v_{\text{air}}$ is a spatially nonlinear function based on the exponential terms in (3). Consequently, the system (8) is also inherently nonlinear and challenging to solve analytically.

It is reasonable to assume that the tracer objects generally follow the airflow [11]. Therefore, we can compute the 2D object velocity from (8) as follows,

$$\mathbf{v}_{\text{obj}} = \max(v_{\text{obj}}, 0) \begin{bmatrix} \cos \alpha \\ \sin \alpha \end{bmatrix} \quad (9)$$

where $\alpha$ is the radial direction from the stagnation point, determining the airflow direction (see Section III-A). The max operator ensures that the object is either stationary or moves in the direction of $\alpha$.

To predict the timing of the object dynamics, we utilized the time delay model developed in our previous work [11], which estimates the time required for object velocity to respond to a newly set air nozzle orientation.

In summary, equations (8) and (9) provide an object motion model for a jet-induced airflow field near a surface. This model also allows for the simulation of object motion, e.g., using the ODE113 solver in MATLAB. Fig. 4 shows both experimental and predicted object trajectories as well as object velocity magnitudes, indicating similar trends and magnitudes. The MAPEs between the experimental and predicted end-of-trajectory positions are 9% and 19.3% for 0° and 22.5° tilt angles, respectively, indicating good alignment.

This level of prediction error is comparable to the inherent stochasticity of such a system. For example, when a tracer object is placed at the same initial position (within 5-mm precision) and its end-of-trajectory position is repeatedly measured at a 0° tilt angle, the MAPE between the first measurement and subsequent ones is 7.7%, reflecting the natural variability in the object motion [11].

### C. Results of System Identification for Irregularly Shaped Objects

We also tested the SINDy algorithm (Section IV-A) on irregularly shaped objects, specifically cotton wads (⌀ ~5 cm, 0.22 g), following the same data collection procedures as for the tracer objects (Appendix D). As a result, we identified a system identical to (8), with coefficients $\xi_1 = -4.25$, $\xi_2 = 3.46$, and $\xi_3 = -4.12$, achieving a goodness of fit of $R^2 = 0.88$. The 2D velocities were computed analogously to the tracer objects from (9). Fig. 5 shows both the experimental and predicted trajectories and velocity magnitudes of the cotton wads. The MAPEs between the experimental and predicted end-of-trajectory positions are 18.4% and 26.9% for 0° and 22.5° tilt angles, respectively, indicating reasonable alignment. These results demonstrate the effectiveness of the approach for irregularly shaped objects, despite the simplicity of the identified system, which uses only three coefficients.

## V. SINGLE- AND MULTI-OBJECT MANIPULATION

The airflow field model (Section III) and the identified object dynamics (Section IV) allow us to design a model-based controller that can achieve single- and multi-object manipulation.

### A. Model-Based Controller

To design the model-based controller, we employed the cross-entropy method (CEM) [22], a gradient-free stochastic optimizer applied in various domains, such as robotics [23], autonomous vehicles [24], and model-based reinforcement learning [25], for motion planning and control.

CEM seeks to find an air nozzle orientation, defined by its pan $\theta$ and tilt $\varphi$ angles (see Fig. 1), that produces object trajectories minimizing the cost function. Rather than directly seeking the nozzle orientation, the algorithm focuses on finding an optimal stagnation point $\mathbf{s}^*$, which can then be converted into the nozzle orientation using (1) and (2). Stagnation points are randomly sampled from a Gaussian distribution, parameterized by its mean $\boldsymbol{\mu}$ and variance $\boldsymbol{\Sigma}$, both of which are iteratively updated by the algorithm.

The algorithm begins by setting the initial values for the mean $\boldsymbol{\mu}_0$ and variance $\boldsymbol{\Sigma}_0$ of the Gaussian distribution. Other initial parameters include the number of detected objects to be manipulated $M$, their initial positions, simulation timespan $\Delta T$, convergence threshold $\boldsymbol{\Sigma}^*$, maximum number of iterations $i_{\max}$, and reference points, defined by the manipulation task.

For each stagnation point sampled from the Gaussian distribution, the corresponding object trajectories are predicted by numerically solving (8) and (9) over the timespan of $\Delta T$. The cost of each sampled stagnation point is then evaluated by comparing the predicted end-of-trajectory positions of the objects with their respective reference points. This comparison uses the following cost function,

$$J = \begin{cases} J_P & \text{if } \exists j \in \{1, \ldots, M\} \text{ such that } \|\mathbf{p}_j^{\text{init}} - \mathbf{s}\| < \delta_{\min} \\ \sum_{j=1}^{M} \|\mathbf{p}_j^{\text{end}} - \mathbf{r}_j\| & \text{otherwise} \end{cases} \quad (10)$$

where $\sum\cdot$ denotes summation, $\mathbf{r}_j$ is the reference point for the object corresponding to its expected end-of-trajectory position $\mathbf{p}_j^{\text{end}}$, and $J_P$ is a penalty applied when the initial position of the object, $\mathbf{p}_j^{\text{init}}$, is excessively close to the stagnation point $\mathbf{s}$, within $\delta_{\min} = 10$ cm. The penalty, set at $J_P = 10^3$, ensures that the objects remain sufficiently distant from $\mathbf{s}$, near which object motion tends to be most stochastic [11]. The sampled stagnation points are then sorted based on cost, and the top $n^*$ samples are selected to update the mean $\boldsymbol{\mu}$ and variance $\boldsymbol{\Sigma}$ of the distribution.

The process continues until either the variance $\boldsymbol{\Sigma}$ falls below the convergence threshold $\boldsymbol{\Sigma}^*$ or the maximum number of iterations $i_{\max}$ is reached. After convergence, the optimal stagnation point $\mathbf{s}^*$ is returned as the mean of the final Gaussian distribution.

Once CEM returns the optimal stagnation point, it is converted into the corresponding air nozzle orientation, generating the airflow field. Subsequently, new positions of the objects are detected using top-view image processing, through the circle detection algorithm [26]. The time lags between changes in the nozzle orientation are determined by $\Delta T$ and a time delay, estimated using our time delay model [11].

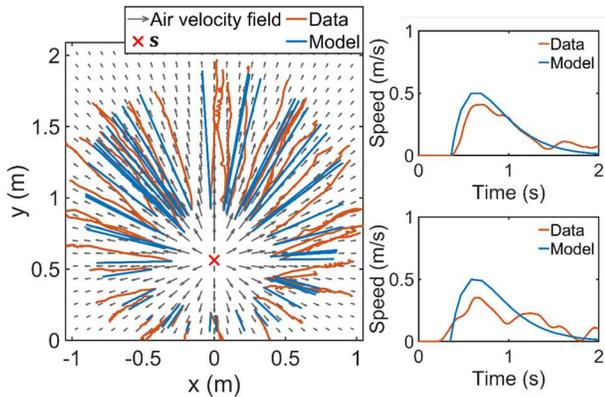

Fig. 5. Experimental (Data) and predicted (Model) trajectories of the irregularly shaped objects (cotton wads), along with the examples of their respective velocity magnitudes (Speed), at $\varphi = 22.5°$ tilt angle of the air nozzle. The stagnation point is denoted by $\mathbf{s}$.

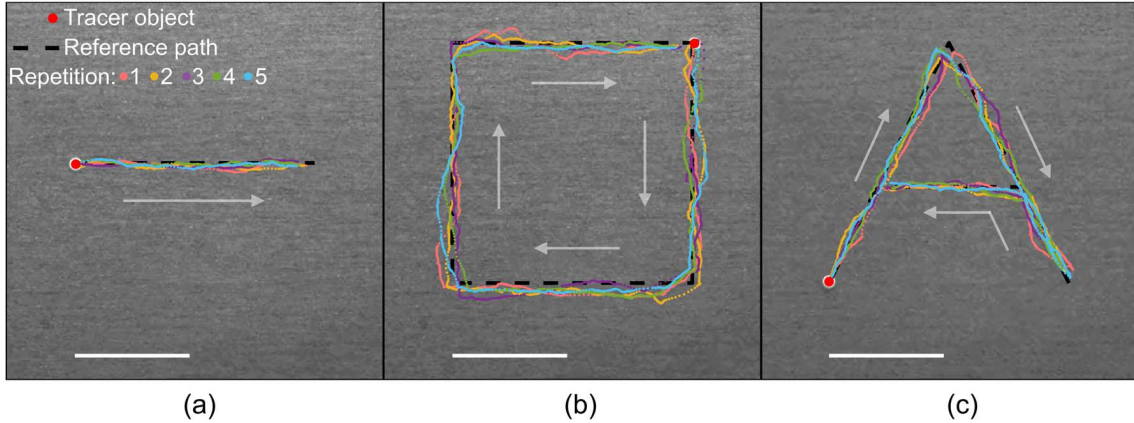

Fig. 6. Results of automatic steering of a tracer object along reference paths with the jet-induced airflow field. The reference paths include (a) a line, (b) a square, and (c) the letter 'A'. The arrows represent the direction of motion of the object. Scale bars: 25 cm.

The initial parameter values of the algorithm were tuned for each manipulation task experimentally (see Sections V-B and V-C for more details). To predict object trajectories by solving (8) and (9), we used the relatively long timespan $\Delta T$ of 1.5 seconds to reduce transient airflow effects. These effects are not captured by airflow model, as it is based on steady-state measurements and simulations (see Appendices A and B).

### B. Single-Object Manipulation

We evaluated CEM (Section V-A) in automatic single-object manipulation through path-following tests with the tracer objects, covering reference paths of varying complexity: a line, a square, and the letter 'A' (see the results in Fig. 6 and supplementary video). According to the path-following errors (Table I, rows 1–3), CEM slightly outperformed the feedback controller from our previous work for the tracer objects [11], achieving a mean maximum error of 4.4 cm—roughly half of the previously achieved 8.89 cm. This highlights the effectiveness of the airflow field model and the identified object dynamics for manipulation.

TABLE I
PATH-FOLLOWING ERRORS AND COMPLETION TIMES

| Path | Mean Error (cm) | Standard Deviation (cm) | Maximum Error (cm) | Completion time (min, s) |
|---|---|---|---|---|
| Line | 0.67 | 0.38 | 2.52 | 38 s |
| Square | 1.24 | 0.86 | 4.40 | 2 min, 34 s |
| 'A' | 0.82 | 0.63 | 3.29 | 1 min, 48 s |
| Octagon | 5.65 | 4.18 | 20.32 | 3 min, 25 s |

To apply CEM, each reference path was divided into reference points with a spacing of 10 cm. The algorithm then steered the object from one reference point to another. When the object was sufficiently close to the current reference point, i.e., within 4 cm in our implementation, the next point was selected. Additionally, since the tracer objects move radially away from the stagnation point, according to (9), the search space for the optimal stagnation point $\mathbf{s}^*$ was limited to the line connecting the object and the reference point. Consequently, stagnation points were sampled from a univariate Gaussian distribution with its mean $\mu \in \mathbb{R}$ and variance $\Sigma \in \mathbb{R}$ corresponding to the distance from the initial position of the object in the direction away from the reference point. The other parameter values were: $\mu_0 = 41.8$ cm, $\Sigma_0 = 20.9$ cm$^2$ (the initial parameters of the distribution), $n = 25$, $n^* = 3$, $i_{\max} = 5$, and $\Sigma^* = 0.3$ cm$^2$ (the convergence threshold).

Note that the manipulation (Fig. 6) was conducted at meter-scale distances, with the distance between the nozzle exit and the object ranging from 1.31 m to 1.52 m and reference path lengths ranging from 52 cm to 209 cm.

### C. Multi-Object Manipulation

We further evaluated the performance of CEM in multi-object manipulation tasks, including automatic path-following, aggregation, and sorting (see the results in Fig. 7 and supplementary video). These tasks involved not only the tracer objects but also the irregularly shaped objects, whose dynamics were identified (see Section IV-C).

In the path-following task shown in Fig. 7(a), four tracer objects were simultaneously steered along an octagonal reference path (see path-following errors in Table I, row 4). The objects moved between the corners of the octagon, with each corner serving as a reference point. The reference point switched when the mean distance between the objects and the current point fell below 6.3 cm. CEM completed the path in 205 seconds, with the maximum pairwise distance between the objects never exceeding 26.1 cm. As a reference for the pairwise distance, we used the 'span', defined as the distance between the first and last objects when tightly arranged in a row. It was calculated as follows: object diameter × (number of objects - 1), which equals 9 cm in this case.

In the aggregation task, shown in Fig. 7(b), 25 tracer objects, initially scattered with a maximum pairwise distance of 1.2 m, were collected into a circular reference zone (ø 0.37 m) within 105 seconds. The center of the reference zone was used as the reference point.

In the sorting task shown in Fig. 7(c), three tracer objects and three cotton wads were sorted into two circular reference zones (ø 0.21 m) based on object type, in 40 seconds. The objects were initially mixed in one group, and the centers of the reference zones, serving as reference points, were 1.25 m apart.

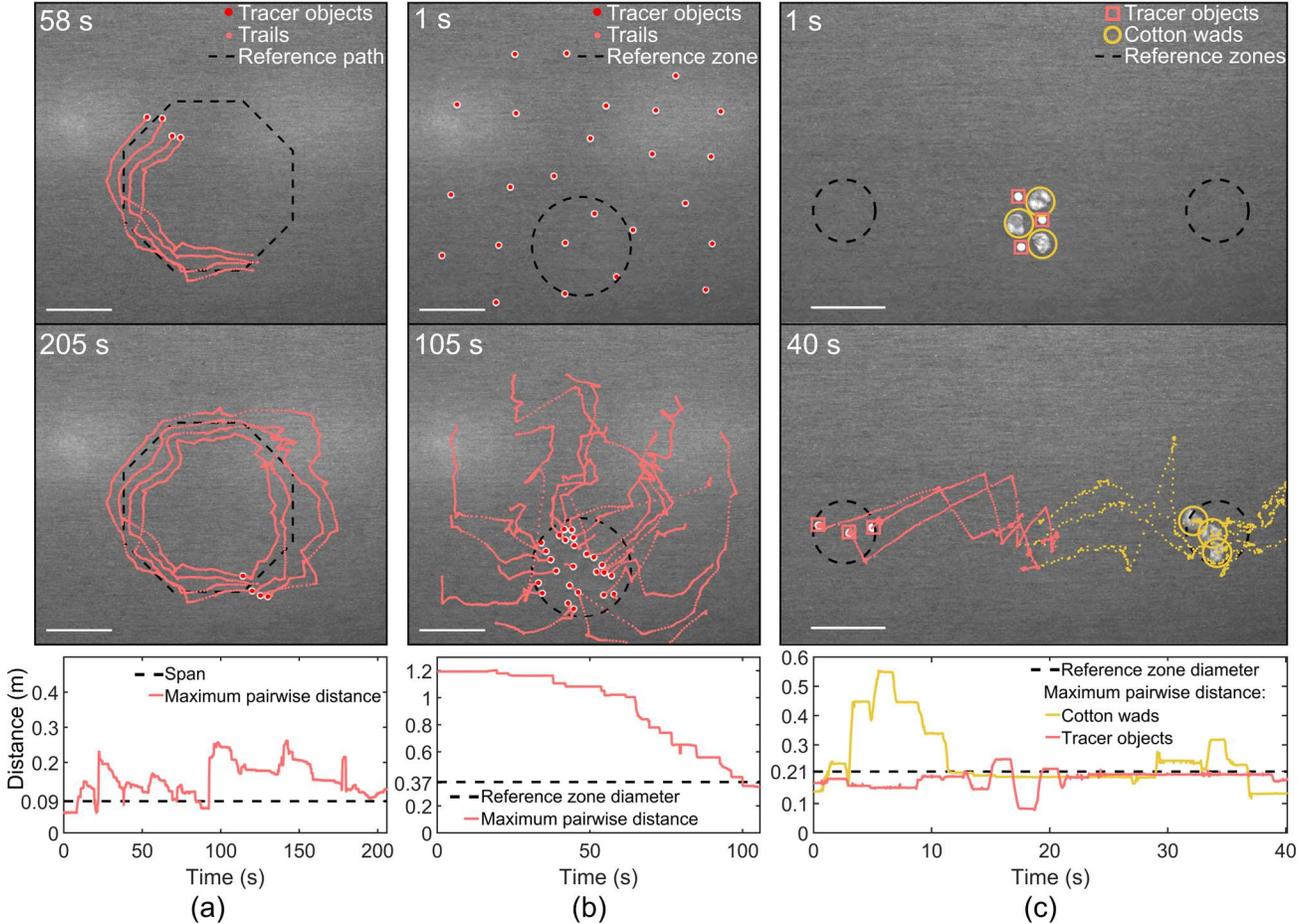

Fig. 7. Results of automatic multi-object manipulation with the jet-induced airflow field: (a) Steering four tracer objects along the reference path; (b) Aggregating 25 tracer objects into the reference zone; (c) Sorting three tracer objects and three cotton wads into two reference zones according to object type. The last row shows the respective dynamics of the maximum pairwise distances between objects. 'Span' (last row, first column) refers to the distance between the first and last objects when tightly arranged in a row. Scale bars: 25 cm.

To apply CEM to the path-following and aggregation tasks, the search space for the optimal stagnation point $\mathbf{s}^*$ was limited to the line connecting the farthest object to the reference point. This heuristic helped reduce the risk of any object exiting the feasible manipulation region, where motion control in any direction over the surface is possible [11]. Consequently, stagnation points were sampled from a univariate Gaussian distribution with its mean $\mu \in \mathbb{R}$ and variance $\Sigma \in \mathbb{R}$ corresponding to the distance from the farthest object in the direction away from the current reference point. The other parameter values were identical to those used in the single-object manipulation (Section V-B).

For the sorting task, stagnation points were sampled from a bivariate Gaussian distribution with $\boldsymbol{\mu} \in \mathbb{R}^2$ and covariance matrix $\boldsymbol{\Sigma} \in \mathbb{R}^{2\times 2}$. The other parameter values were: $\boldsymbol{\Sigma}_0 = \text{diag}(31.3, 31.3)$ cm$^2$, $n = 100$, $n^* = 10$, $i_{\max} = 5$, and $\boldsymbol{\mu}_0 = \text{mean}(\mathbf{P}^{\text{init}})$ cm, where $\text{mean}(\mathbf{P}^{\text{init}})$ is the 2D mean of the initial positions of the objects. For the convergence threshold $\boldsymbol{\Sigma}^*$, we used the Frobenius norm of the covariance matrix, set to $\|\boldsymbol{\Sigma}\|_F = 0.5$ cm$^2$.

## VI. Conclusion

This article presented an approach to analytically modelling the jet-induced airflow field and identifying object dynamics within that field for automatic and remote single- and multi-object manipulation. The analytical model was developed by capturing the key properties of the jet-induced airflow field. Using model-based air velocity predictions, the SINDy algorithm identified the ODE governing the object dynamics. For object motion control, we employed CEM, demonstrating its effectiveness across various manipulation tasks. The proposed approach consolidates our explorative study of the airflow-field-based manipulation method [11], providing new insights into both the airflow field and object dynamics within it, and extending the capability to multi-object manipulation.

## Appendix A: Air Velocity Measurements

Measurements of air velocity magnitude were conducted using a hot-wire anemometer (degreeC UAS150) at equally spaced points on an 11 by 11 grid, covering a 2×2-m workspace, at 1.5 cm away from the surface. The measurement for each point was averaged over a one-minute period. Before each measurement was taken, the jet-induced airflow field was allowed to reach its steady state. According to the CFD simulations (Appendix B), the steady state is reached in approximately 8 seconds. Measurements were taken at three nozzle tilt angles: 0°, 22.5°, and 45°. The flowrate was set to 142 L/min, with other experimental conditions matching those in Section II-A. To spatially interpolate the measurements over the surface, we used Makima interpolation [16].

## APPENDIX B: CFD SIMULATIONS

CFD simulations of the jet-induced airflow field were performed on COMSOL Multiphysics 6.0 using a single-phase turbulent flow module and a standard k–ε interface, which solved the Reynolds-averaged Navier–Stokes equations in three dimensions. The simulations were repeated for three tilt angles of the nozzle, 0°, 22.5°, and 45°, covering a workspace of over 2×2 m. Other simulation conditions matched those used in the measurements (Appendix A). Each simulation ran until a steady state was reached at 8 seconds, after which air velocity fields 1.5 cm above the surface were extracted and averaged.

## APPENDIX C: DATA FUSION

We considered the air velocity measurements (Appendix A) conducted near the stagnation points to be unreliable due to high turbulence. Specifically, turbulence intensity near the stagnation points can reach approximately 10% [11], causing rapid and high-amplitude fluctuations in velocity, which are challenging to measure with the hot-wire anemometer. On the other hand, the CFD simulations, described Appendix B, accurately predicted the air velocity to be zero or very low at the stagnation points, which agrees with previous studies on impinging jet airflows [12], [15]. Thus, to improve the accuracy of the airflow field model, we fused the simulation data with the measurements by applying a distance-based scaling factor. This factor maximized the weight of the simulation data at the stagnation point, gradually transitioning to relying more heavily on the experimental data as the distance from this point increased. The fused data was then used to fit the air velocity magnitudes (3).

## APPENDIX D: OBJECT-TRACKING DATA

We tracked object trajectories for two nozzle tilt angles, 0° and 22.5°, with 50 trajectories per angle. Objects were manually placed at distances from 20 cm to 58 cm from the stagnation point. At these distances, static pressure remains low [11], but air velocity is sufficient to set the objects in motion.

The data collection procedure consisted of three main steps: (i) placing an object on the surface with the nozzle in an idle orientation ($\theta = -90°, \varphi = 50°$) to keep the object stationary; (ii) emitting the air jet and waiting 8 seconds for the airflow to reach its steady state; and (iii) adjusting the nozzle to the desired orientation to record the object trajectory using a top-view camera. The trajectories were recorded for two seconds, allowing the object velocity to reduce to zero or negligible levels (see Fig. 4 and 5).

Following the system identification process (Section IV), we collected velocity data, $\mathbf{V}_{\text{obj}}^{k}$ and $\mathbf{V}_{\text{obj}}^{(k+1)}$ in (6), from the recorded trajectories. The velocity magnitudes were measured using the central finite difference approximation, given by $v_{\text{obj}}^{k} = \|\frac{\mathbf{p}^{(k+1)} - \mathbf{p}^{(k-1)}}{2\Delta t}\|$, where $\mathbf{p}^{(k+1)}$ and $\mathbf{p}^{(k-1)}$ are the 2D positions of the object at timesteps $k+1$ and $k-1$ respectively, and $\Delta t$ is the time interval between the timesteps. With a frame rate of 40 Hz, $\Delta t$ was 25 milliseconds. To simplify the system identification process, datapoints where the objects remained stationary were discarded, as the dry friction force, $\mathbf{F}_{\text{f}}$ in (4), may be discontinuous at zero velocity [18].

## ACKNOWLEDGMENT

This work was supported by the Aalto Doctoral School of Electrical Engineering. The authors thank Dr. Houari Bettahar for commenting on the manuscript.

The authors declare that they have no competing interests.